\documentclass[runningheads]{llncs}
\usepackage[T1]{fontenc}
\usepackage{graphicx}
\usepackage{amsmath}
\usepackage{amssymb}
\usepackage{booktabs}
\usepackage{makecell}

\begin{document}
\title{PokeFusion Attention: A Lightweight Cross-Attention Mechanism for Style-Conditioned Image Generation}
\titlerunning{PokeFusion Attention for Style-Conditioned Generation}
\author{Jingbang Tang}

\authorrunning{J. Tang}

\institute{
Independent Researcher, Vancouver, Canada\\
Formerly with School of Computing, Universiti Kebangsaan Malaysia (UKM), Malaysia\\
\email{tjbleo616@gmail.com}
}
\maketitle              % typeset the header of the contribution
\begin{abstract}
Style-conditioned text-to-image (T2I) generation with diffusion models requires both stable character structure and consistent, fine-grained style expression across diverse prompts. Existing approaches either rely on text-only prompting, which is often insufficient to specify visual style, or introduce reference-based adapters that depend on external images at inference time, increasing system complexity and limiting deployment flexibility.

We propose PokeFusion Attention, a lightweight decoder-level cross-attention mechanism that models style as a learned distributional prior rather than instance-level conditioning. The method integrates textual semantics with learned style embeddings directly within the diffusion decoder, enabling effective stylized generation without requiring reference images at inference time. Only the cross-attention layers and a compact style projection module are trained, while the pretrained diffusion backbone remains frozen, resulting in a parameter-efficient and plug-and-play design.

Experiments on a stylized character generation benchmark demonstrate that the proposed method improves style fidelity, semantic alignment, and structural consistency compared with representative adapter-based baselines, while maintaining low parameter overhead and simple inference.
\end{abstract}

\section{Introduction}

Style-conditioned text-to-image (T2I)~\cite{shen2024imagpose} generation with diffusion models has made rapid progress, yet it remains fragile in character-centric settings. Unlike general T2I synthesis, character generation requires both consistent global structure and stable, fine-grained style expression across diverse prompts. In practice, text-only prompting is often insufficient to specify such visual details, leading to style drift and geometric inconsistency.

Existing approaches typically address this issue either by introducing reference images at inference time or by injecting additional conditioning branches. However, such designs increase system complexity and create dependency on external inputs, limiting deployment flexibility. This raises a key question: can style be internalized as a reusable representation, enabling controllable generation without relying on explicit reference inputs at inference time?

A common remedy is to introduce reference images at inference time, e.g., by conditioning on exemplar style/appearance. While effective, reference-based pipelines add extra user burden, increase system complexity, and create a hard dependency on external inputs during deployment. This limitation becomes more pronounced in interactive editing and controllable generation scenarios, where users expect lightweight, modular control without repeatedly supplying reference images. Recent controllable diffusion research therefore trends toward parameter-efficient and composable control modules that adapt large pretrained backbones without full finetuning~\cite{hu2022lora}. This design philosophy has been repeatedly validated in broader controllable generation and editing tasks~\cite{ruiz2023dreambooth,tumanyan2023pnp,hertz2024stylealigned}.

Existing controllability methods for diffusion models largely follow two families. The first family, exemplified by ControlNet-style designs, injects structural cues (e.g., edges, poses, depth) into the U-Net to enforce spatial constraints. These methods excel at geometry control but are less suitable for high-level ``style as a distribution'' control, where style is modeled as a global rendering prior rather than spatial constraints~\cite{karras2020stylegan}. The second family leverages cross-attention adapters (e.g., IP-Adapter-like branches) to import rich style signals from an image encoder. Although powerful, these approaches typically require reference images at inference time and often introduce additional branches and feature pathways, which increases memory/latency and complicates portability across diffusion backbones.

To address this, we propose \textbf{PokeFusion Attention}, a lightweight decoder-level cross-attention mechanism that models style as a learned distributional prior rather than instance-level conditioning. The key idea is to learn a compact style embedding space during training and fuse it with textual semantics \emph{within the diffusion decoder}, where semantic-to-visual binding is naturally mediated by cross-attention. 

Decoder cross-attention provides a suitable interface for injecting style information, as it directly influences how semantic tokens are translated into visual patterns. Instead of introducing additional network branches or relying on external images at inference time, PokeFusion adopts a dual-branch attention design that jointly attends to (i) text tokens for content and (ii) learned style tokens for rendering behavior. 

We freeze the pretrained backbone and update only the cross-attention layers together with a lightweight style projection module, resulting in a parameter-efficient and plug-and-play controller that can be readily integrated into existing diffusion pipelines.

\begin{figure}[t]
    \centering
    \includegraphics[width=\linewidth]{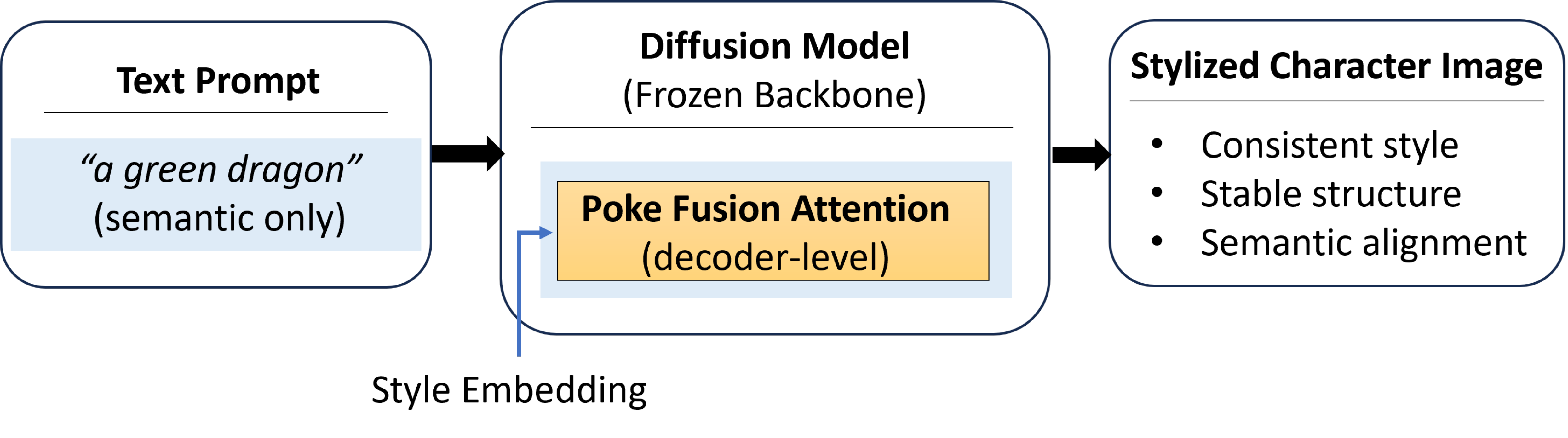}
    \caption{
    Training and inference overview of style-conditioned character generation. 
    During training, style images are used to learn a compact style embedding. 
    At inference time, generation is conditioned only on text, without requiring reference images.
    }
    \label{fig:task_overview}
\end{figure}

As illustrated in Fig.~\ref{fig:task_overview}, we consider a reference-free style-conditioned character generation setting, where only a text prompt is provided at inference time and no reference images are available for guidance. Under this setting, the key challenges are to mitigate style drift across diverse prompts while preserving stable character structure and accurate semantic alignment.

\paragraph{Contributions.}
Our main contributions are threefold:
\begin{itemize}
    \item We introduce a decoder-level cross-attention mechanism that models style as a learned distributional prior, enabling controllable stylized generation without reliance on instance-level reference inputs at inference time.
    \item The method updates only cross-attention layers and a lightweight style projection module, enabling parameter-efficient and plug-and-play adaptation.
    \item On a Pok\'emon-style character generation dataset, we demonstrate improved style fidelity, semantic alignment, and shape/appearance consistency compared with representative adapter-based baselines.
\end{itemize}

\section{Related Work}
\label{sec:rw}

\subsection{Diffusion-Based Text-to-Image Generation and Controllability}

Diffusion models have become the dominant paradigm for text-to-image generation, with representative systems such as DALL-E~2~\cite{ramesh2022dalle2}, Imagen~\cite{saharia2022imagen}, and Stable Diffusion~\cite{rombach2022ldm}. By leveraging large-scale pretrained text encoders, these models achieve strong semantic alignment between prompts and synthesized images, and latent diffusion further improves efficiency by performing denoising in a compressed latent space conditioned on a frozen CLIP text encoder~\cite{rombach2022ldm}. Despite this progress, most mainstream pipelines remain primarily text-conditioned, which is often insufficient for character-centric generation that demands fine-grained style expression and stable shape/appearance. In practice, text prompts alone frequently lead to style drift and structural inconsistency, motivating additional conditioning and adaptation mechanisms beyond fixed textual encodings~\cite{liu2022ptuningv2,houlsby2019adapter}.

\subsection{Control Modules}

To enhance controllability, prior works introduce auxiliary control modules for diffusion models. ControlNet~\cite{zhang2023controlnet} and T2I-Adapter~\cite{mou2023t2iadapter} inject structural cues (e.g., edges, depth, poses) to enforce spatial constraints and improve geometry/layout adherence, but their primary focus is structure-guided generation rather than high-level style consistency or character appearance stability. Another line of research employs adapter-based or cross-attention-based designs to inject style information through additional visual or multimodal representations. IP-Adapter~\cite{ye2023ipadapter}, for example, introduces a parallel image-encoding branch and fuses visual features into attention layers for reference-based style transfer; however, such methods typically require reference images at inference time, increasing architectural complexity and creating deployment-time dependency. Moreover, conflicts between text and visual conditions may cause overfitting or identity/appearance inconsistency~\cite{gal2023textualinversion}.

In contrast, recent evidence suggests that modifying decoder-level attention can provide parameter-efficient controllability while preserving the pretrained backbone~\cite{mokady2023nulltext}. Conceptually, IP-Adapter conditions generation on instance-level visual exemplars by mapping a specific reference image to style features, whereas PokeFusion Attention learns a global style prior that is internalized within decoder cross-attention layers. This allows style control without requiring reference images at inference time, and avoids the additional branches and feature pathways used in prior adapter-based methods. Following this direction, our work targets \emph{reference-free} style-conditioned generation by introducing a lightweight decoder-level cross-attention fusion mechanism. By adapting only cross-attention layers and a lightweight style projection module, our approach avoids network duplication and eliminates inference-time reliance on external images, enabling efficient and portable stylized character generation.

\section{Proposed Method}
\label{sec:method}

\subsection{Overview}

We consider a standard text-to-image diffusion model that generates an image from a text prompt $\mathbf{c}$. Let $\mathbf{h} \in \mathbb{R}^{L \times d}$ denote the hidden representation of the U-Net decoder at a given layer. In existing diffusion models, generation is primarily conditioned on textual embeddings, which limits fine-grained style control and often results in unstable character appearance in character-centric scenarios.

\begin{figure*}[t]
    \centering
    \includegraphics[width=\linewidth]{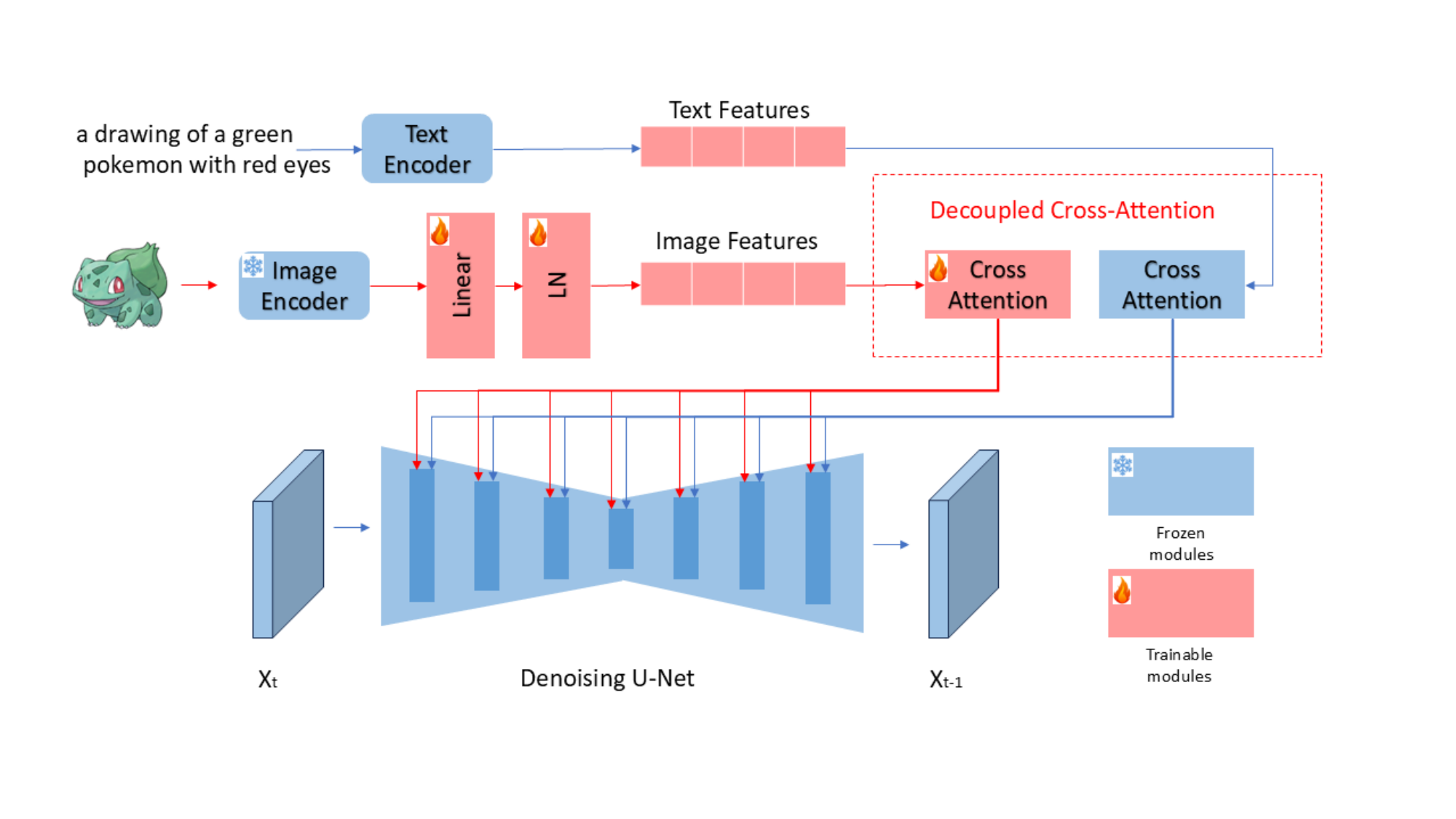}
    \caption{Architecture of PokeFusion Attention. Only the cross-attention layers and the style projection module (Linear + LayerNorm) are trainable (highlighted in red), while the rest of the diffusion U-Net remains frozen.}
    \label{fig:pokefusion_arch}
\end{figure*}

We propose \textbf{PokeFusion Attention}, a lightweight decoder-level cross-attention mechanism for \emph{style-conditioned generation without requiring reference images at inference time}. As illustrated in Fig.~\ref{fig:pokefusion_arch}, our key idea is to decouple textual and style conditioning into two parallel cross-attention branches within the decoder, and fuse them through a simple yet effective weighting scheme. Unlike prior approaches that introduce additional encoder branches or require reference images at inference time, PokeFusion Attention modifies only decoder cross-attention layers. During training, only these attention parameters are updated, while the diffusion backbone remains frozen, enabling efficient and portable adaptation. Note that only the cross-attention layers and the style projection module are trainable, rather than the entire decoder. Importantly, although style images are used during training to learn a global style representation, no reference images are required at inference time.

\subsection{Style Feature Projection}

To represent visual style information, we associate each training image with a style embedding. Let $\mathbf{I}_s$ denote a style image sampled from a style-specific dataset, and let $\mathbf{x} \in \mathbb{R}^{d_x}$ be its visual feature extracted by an image encoder $\phi(\cdot)$. We project $\mathbf{x}$ into a style embedding $\mathbf{s} \in \mathbb{R}^{d}$ using a linear projection followed by normalization:
\begin{equation}
\mathbf{s} = \mathrm{LayerNorm}(W\mathbf{x} + \mathbf{b}),
\label{eq:style_proj}
\end{equation}
where $W$ and $\mathbf{b}$ are trainable parameters. This projection aligns visual style features with the text embedding space used by the diffusion model. Importantly, the resulting style embedding is not tied to individual reference images, but captures global rendering statistics of a target style domain, enabling style control as a distribution-level prior. In our experiments, a single learned style embedding representing the Pok\'emon-style domain is used during inference.

\subsection{Decoder-Level Cross-Attention Fusion}

Let $\mathbf{t} \in \mathbb{R}^{M \times d}$ denote text token embeddings. At each decoder block, we compute two cross-attention outputs conditioned on text and style, respectively:
\begin{align}
\mathbf{A}_{\text{text}} &= \mathrm{Attn}(\mathbf{h}, \mathbf{t}), \\
\mathbf{A}_{\text{style}} &= \mathrm{Attn}(\mathbf{h}, \mathbf{s}),
\end{align}
where $\mathrm{Attn}(\cdot)$ denotes standard scaled dot-product cross-attention. The two branches are fused as:
\begin{equation}
\mathbf{A}_{\text{fused}} = (1-\alpha)\mathbf{A}_{\text{text}} + \alpha \mathbf{A}_{\text{style}},
\label{eq:fusion}
\end{equation}
where $\alpha \in [0,1]$ controls the contribution of style information. This fusion operation is applied uniformly to all decoder cross-attention layers, enabling consistent style modulation throughout the denoising process. In practice, $\alpha$ is fixed across layers and timesteps during inference unless otherwise specified.

\subsection{Training Objective}

We adopt the standard denoising diffusion training objective. Given a noisy latent $\mathbf{y}_t$ at timestep $t$, the model predicts the added noise $\hat{\boldsymbol{\epsilon}}_\theta$ conditioned on text and style:
\begin{equation}
\mathcal{L} = \mathbb{E}\left[\left\|\boldsymbol{\epsilon} - \hat{\boldsymbol{\epsilon}}_\theta(\mathbf{y}_t, t, \mathbf{c}, \mathbf{s})\right\|_2^2\right].
\end{equation}
During training, text or style conditioning is randomly dropped to enable classifier-free guidance. At inference time, a guidance scale $\omega$ balances conditional and unconditional predictions without introducing additional inputs.

\subsection{Relation to Prior Work}

Compared to ControlNet and T2I-Adapter, which inject external structural features or auxiliary control signals through additional pathways, our method targets high-level style conditioning without modifying encoder components. Unlike IP-Adapter, which relies on reference images and separate encoding branches at inference time, PokeFusion Attention achieves style-conditioned generation through decoder-only cross-attention adaptation. This design enables lightweight, reference-free inference while preserving the original diffusion backbone.

\section{Experiments and Analysis}
\label{sec:exp}

All evaluations are conducted without providing reference images at inference time. We evaluate \textbf{PokeFusion Attention} on a stylized character generation task and compare it with representative controllable diffusion baselines, as described in Section~\ref{sec:exp}. Our experiments aim to answer the following questions: (i) whether decoder-level style fusion improves style fidelity and semantic alignment, (ii) how the proposed method compares with reference-based adapters, and (iii) which components contribute most to the observed performance gains. 

Overall, the experimental results demonstrate that decoder-level style fusion consistently improves both style fidelity and semantic alignment across diverse prompts. Compared with text-only prompting, \textbf{PokeFusion Attention} reduces style drift while maintaining stable character structure. 

Despite operating without reference images at inference time, the proposed method achieves performance comparable to or better than reference-based adapters across multiple evaluation metrics, suggesting that explicit visual references are not strictly required when style information is effectively integrated at the decoder level. 

Ablation studies further show that cross-attention-based style injection is a key contributor to the observed performance gains, as removing this component leads to noticeable degradation in both quantitative metrics and visual quality. 

We do not explicitly compare different injection locations, and instead focus on evaluating the effectiveness of decoder-level style fusion as a simple and practical design choice.

\subsection{Dataset}

\begin{figure}[t]
    \centering
    \includegraphics[width=0.9\linewidth]{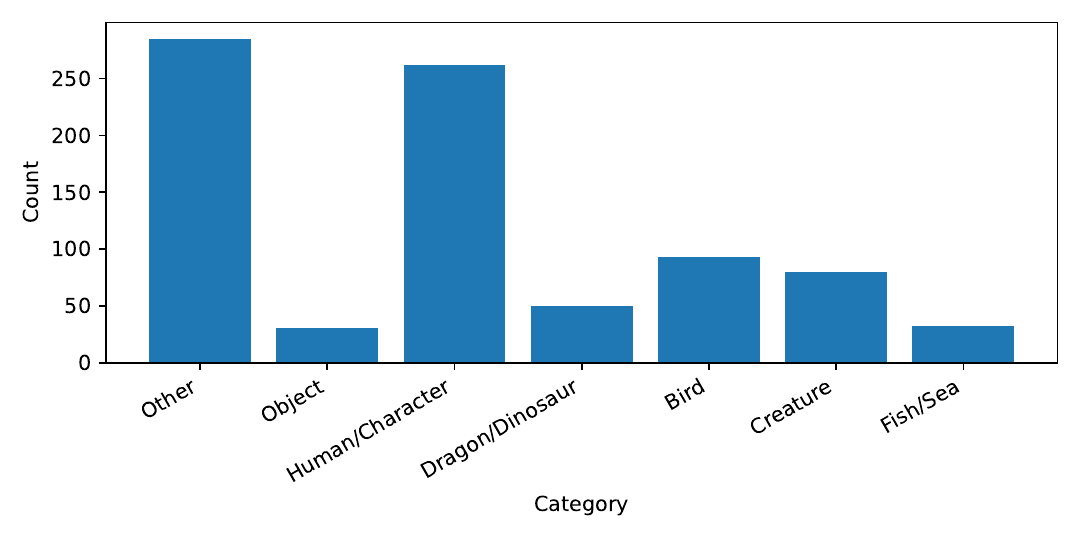}
    \caption{Distribution of subject categories in the \textit{pokemon-blip-captions} dataset based on caption keyword analysis.}
    \label{fig:dataset_category}
\end{figure}

We conduct experiments on the \textit{pokemon-blip-captions} dataset, which consists of 833 Pok\'emon-style character images paired with descriptive text prompts. The captions describe fine-grained visual attributes such as shape, pose, and color. The dataset provides a controlled benchmark for character-centric generation, where preserving global structure and style consistency is critical. All images are resized to $256 \times 256$ following the backbone's native resolution for training and evaluation. Fig.~\ref{fig:dataset_category} summarizes the subject category distribution inferred from caption keywords, indicating a balanced coverage of common character types.

\subsection{Baselines}

We compare our method with representative controllable diffusion approaches covering both structure-guided and style-conditioned paradigms.

\textbf{ControlNet Shuffle}~\cite{zhang2023controlnet} injects structural signals to guide generation and is widely used for geometry control. 
\textbf{T2I-Adapter (Style)}~\cite{mou2023t2iadapter} introduces lightweight adapters for conditioning diffusion models with external signals. 
\textbf{Uni-ControlNet} extends structural control to multiple conditions within a unified framework. 
\textbf{IP-Adapter}~\cite{ye2023ipadapter} incorporates image-based conditioning via cross-attention and represents a strong reference-based baseline for style control.

These baselines are selected because they represent dominant design strategies for controllable diffusion, including structural conditioning and reference-based style adaptation.
All methods are evaluated using the same pretrained Stable Diffusion backbone for fair comparison.

\subsection{Evaluation Metrics}

We evaluate generation quality using CLIP-based metrics for semantic alignment and visual consistency. 
We adopt \textbf{CLIP Score}~\cite{hessel2021clipscore} to measure text--image alignment (CLIPScore-T), 
and report image--image similarity in the CLIP embedding space (CLIPScore-I) to assess appearance and style consistency. 

For method comparison with existing baselines (Table~\ref{tab:method_comparison}), we use the standard CLIPScore formulation. 
For ablation studies (Table~\ref{tab:ablation}), we report cosine similarity between CLIP embeddings (scaled by 100 for readability) 
to better capture relative effects of localized architectural changes. 

For clarity, CLIPScore-based metrics (CLIPScore-T / CLIPScore-I) and cosine-similarity-based metrics (CLIP-T / CLIP-I) are reported separately and are not directly comparable in scale.

\subsection{Implementation Details}

All methods are trained on the same dataset using identical diffusion backbones. 
We use the AdamW optimizer with a learning rate of $1 \times 10^{-4}$ and a batch size of 8. 
Training is performed with mixed precision (fp16) and classifier-free guidance. 
Only the cross-attention layers and the style projection module are updated, while all other parameters of the diffusion U-Net remain frozen.
All experiments are conducted using the same inference settings unless otherwise specified.

\subsection{Quantitative Results}

Table~\ref{tab:method_comparison} summarizes the quantitative comparison with the above baselines. 
PokeFusion Attention achieves the best CLIPScore-T and CLIPScore-I scores among adapter-based approaches, 
indicating improved semantic alignment and style fidelity. 
Notably, our method outperforms IP-Adapter while maintaining comparable parameter counts 
and without requiring reference images during inference. 
These results demonstrate that decoder-level style fusion provides an effective and lightweight alternative 
to reference-based adapters.

\begin{table}[t]
\centering
\caption{Comparison of text-to-image generation methods. CLIPScore-T and CLIPScore-I denote CLIPScore-based metrics for text--image and image--image alignment, respectively.}
\label{tab:method_comparison}
\setlength{\tabcolsep}{3pt}
\renewcommand{\arraystretch}{1.0}
\resizebox{\linewidth}{!}{%
\begin{tabular}{lccccc}
\toprule
\textbf{Method} & Reusable & Multi-Prompt & Params (M) & \textbf{CLIPScore-T} & \textbf{CLIPScore-I} \\
\midrule
\multicolumn{6}{l}{\textit{Adapter-Based Methods}} \\
\midrule
ControlNet Shuffle         & Yes & Yes & 361  & 0.432 & 0.618 \\
T2I-Adapter (Style)        & Yes & Yes & 39   & 0.492 & 0.662 \\
Uni-ControlNet             & Yes & Yes & 47   & 0.510 & 0.738 \\
IP-Adapter                 & Yes & Yes & 22   & 0.589 & 0.824 \\
\textbf{PokeFusion (Ours)} & Yes & Yes & 22   & \textbf{0.605} & \textbf{0.839} \\
\midrule
\multicolumn{6}{l}{\textit{Fine-Tuned Models}} \\
\midrule
SD Image Variations        & No  & No  & 860  & 0.550 & 0.768 \\
SD unCLIP                  & No  & No  & 870  & 0.576 & 0.798 \\
\midrule
\multicolumn{6}{l}{\textit{Training from Scratch}} \\
\midrule
\textbf{Open unCLIP}       & No  & No  & 893  & \textbf{0.610} & \textbf{0.855} \\
Kandinsky-2.1              & No  & No  & 1229 & 0.596 & 0.852 \\
Versatile Diffusion        & No  & Yes & 860  & 0.580 & 0.827 \\
\bottomrule
\end{tabular}%
}
\end{table}

\subsection{Qualitative Comparison}

\begin{figure*}[t]
    \centering
    \includegraphics[width=0.9\textwidth]{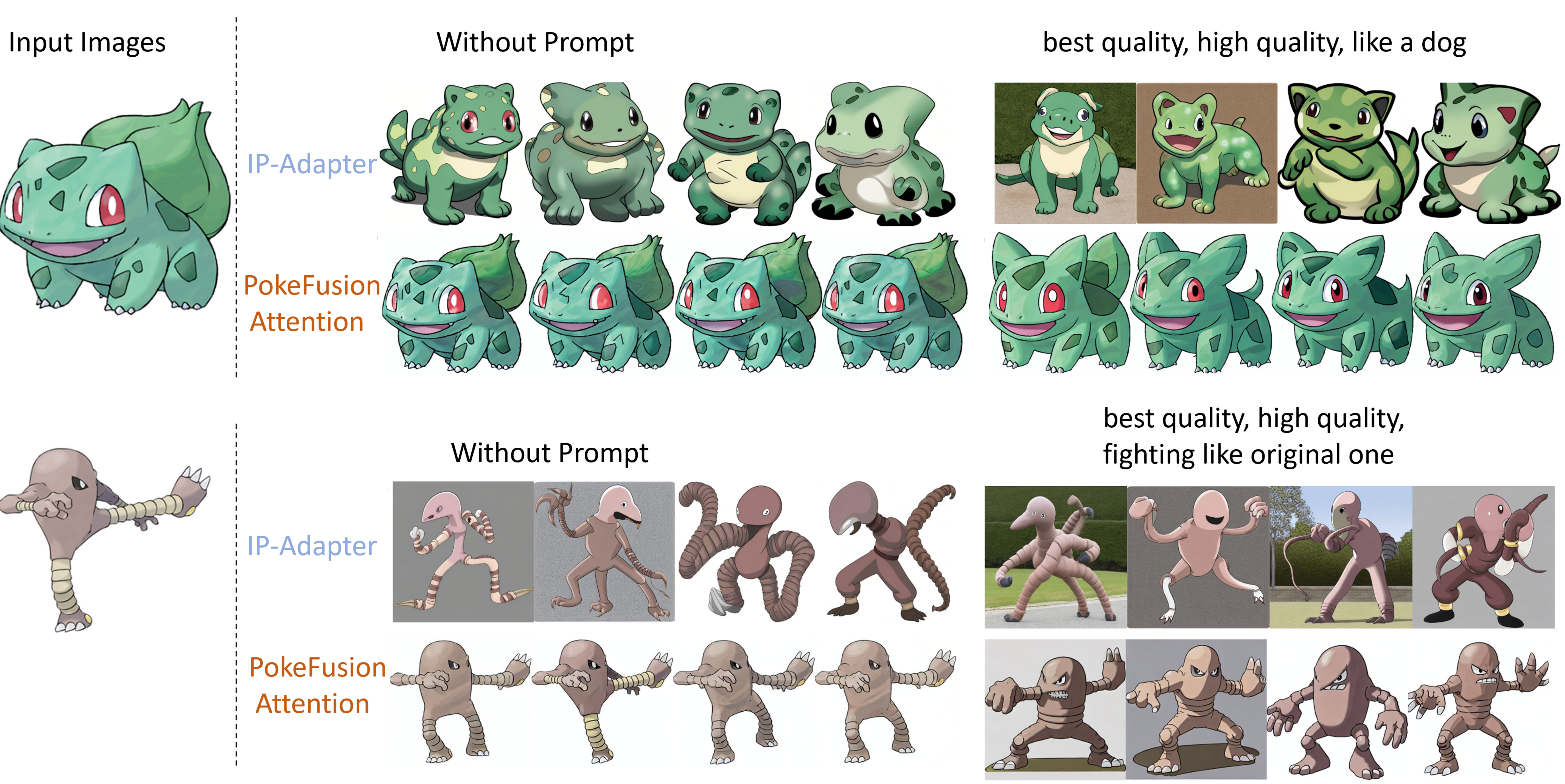}
    \caption{Qualitative comparison between IP-Adapter and PokeFusion Attention.}
    \label{fig:qualitative_comparison}
\end{figure*}

Figure~\ref{fig:qualitative_comparison} presents qualitative comparisons between IP-Adapter and PokeFusion Attention under identical prompts. 
Our method generates characters with more consistent shapes and clearer style expression across different prompts. 
In contrast, IP-Adapter often exhibits noticeable structural variation when reference cues are weak or absent, 
highlighting the advantage of style conditioning without requiring reference images at inference time.

\subsection{Robustness under Inference Settings}

\begin{figure*}[t]
    \centering
    \includegraphics[width=0.9\linewidth]{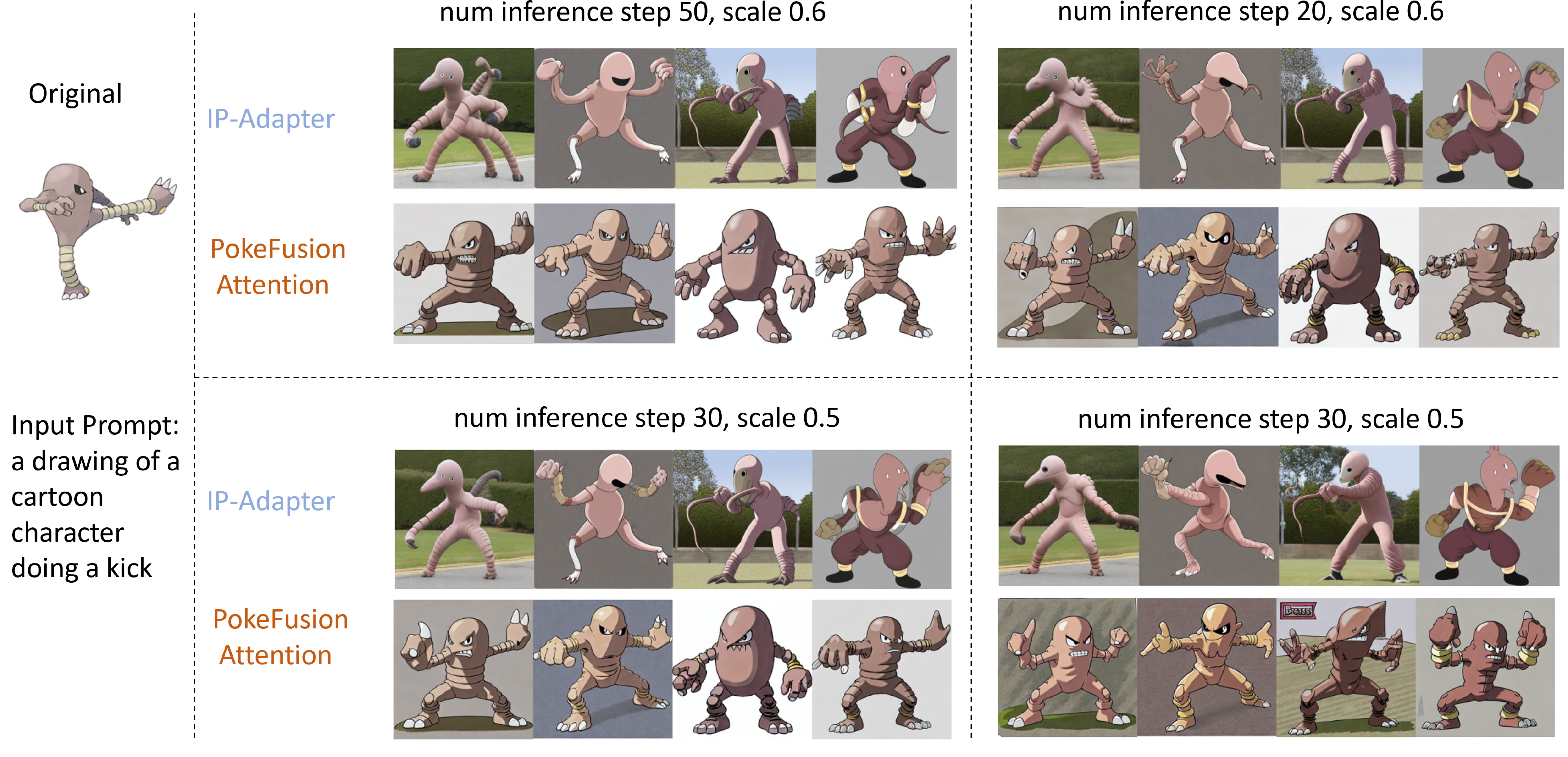}
    \caption{Robustness under varying inference settings.}
    \label{fig:robustness_inference}
\end{figure*}

We further analyze robustness under varying inference settings. 
As shown in Fig.~\ref{fig:robustness_inference}, PokeFusion Attention maintains stable structure and style 
across different sampling steps and guidance scales, whereas IP-Adapter exhibits increased variability. 
This indicates that decoder-level style fusion is less sensitive to inference hyperparameters 
and provides more robust generation behavior.

\subsection{Ablation Study}

We conduct ablation experiments to analyze the contribution of key components in PokeFusion Attention. 
Table~\ref{tab:ablation} summarizes the ablation results.

\begin{table}[t]
\centering
\caption{Ablation study. CLIP-T and CLIP-I denote cosine similarity metrics in the CLIP embedding space.}
\label{tab:ablation}
\small
\setlength{\tabcolsep}{3.5pt}
\begin{tabular*}{\columnwidth}{@{\extracolsep{\fill}} lccc}
\toprule
Method Variant & \makecell[c]{CLIP-T\\(cos)$\downarrow$} & \makecell[c]{CLIP-I\\(cos)$\uparrow$} & \makecell[c]{Style\\Consistency$\uparrow$} \\
\midrule
Text-only (Baseline)           & 28.5 & 60.5 & 62.9 \\
+ Style Embedding (No Fusion)  & 26.1 & 74.7 & 72.4 \\
+ Cross-Attn (Decoder, Frozen) & 25.3 & 80.2 & 75.2 \\
+ Cross-Attn (Decoder, Ours)   & \textbf{24.1} & \textbf{86.9} & \textbf{81.7} \\
\bottomrule
\end{tabular*}
\end{table}

\section{Conclusion}
\label{sec:con}

We show that style-conditioned text-to-image generation without requiring reference images at inference time can be effectively achieved with \textbf{PokeFusion Attention}, a lightweight decoder-level cross-attention mechanism that integrates learned style embeddings with textual semantics while keeping the diffusion backbone frozen. Experiments on the \textit{pokemon-blip-captions} dataset demonstrate consistent improvements in style fidelity, semantic alignment, and generation stability over representative adapter-based baselines, while using the same parameter budget and requiring no reference images at inference time. These results suggest that selectively adapting decoder cross-attention layers is sufficient for internalizing style as a distribution-level prior. Overall, PokeFusion Attention offers an efficient and portable solution for character-centric and domain-specific stylized generation. Future work will extend this framework to other generative tasks, including video generation and multi-style conditioning.

\paragraph{Future Work.}
In future work, we plan to extend PokeFusion Attention to support explicit reference image conditioning during inference without retraining. Recent works have explored hybrid approaches combining reference and prompt conditioning~\cite{hertz2023prompt2prompt}, which may inspire future directions. We also intend to explore multi-style fusion strategies, enabling the model to blend multiple visual identities in a controllable manner. Multi-style diffusion is also gaining traction, particularly in controllable content creation~\cite{ho2022cfg}. Furthermore, integrating dynamic prompt-aware fusion weights and expanding evaluation to additional datasets and user studies will help validate the method’s generality and real-world applicability. Ultimately, our goal is to create a modular, scalable framework for fine-grained, style-aware generation across diverse domains. A modular, scalable architecture aligns with emerging trends in compositional generation frameworks~\cite{liu2022composable}. Recent studies on reference-free and multi-style generation further suggest that distribution-level style control can generalize across domains.

\begin{thebibliography}{00}

\bibitem{shen2024imagpose}
Shen, F., Tang, J.: IMAGPose: A unified conditional framework for pose-guided person generation. OpenReview (2024).

\bibitem{hu2022lora}
Hu, E.J., Shen, Y., Wallis, P., Allen-Zhu, Z., Li, Y., Wang, L., Chen, W.: LoRA: Low-rank adaptation of large language models. In: ICLR (2022)

\bibitem{saharia2022imagen}
Saharia, C., Chan, W., Saxena, S., Li, L., Whang, J., Salimans, T., Ho, J., Fleet, D., Norouzi, M.: Photorealistic text-to-image diffusion models with deep language understanding. In: NeurIPS (2022)

\bibitem{karras2020stylegan}
Karras, T., Aittala, M., Laine, S., Härkönen, E., Hellsten, J., Lehtinen, J., Aila, T.: Analyzing and improving the image quality of StyleGAN. In: CVPR (2020)

\bibitem{ramesh2022dalle2}
Ramesh, A., Dhariwal, P., Nichol, A., Chu, C., Chen, M.: Hierarchical text-conditional image generation with CLIP latents. arXiv preprint arXiv:2204.06125 (2022)

\bibitem{rombach2022ldm}
Rombach, R., Blattmann, A., Lorenz, D., Esser, P., Ommer, B.: High-resolution image synthesis with latent diffusion models. In: CVPR (2022)

\bibitem{liu2022ptuningv2}
Liu, X., Ji, K., Fu, Y., Tam, W., Du, Z., Yang, Z., Tang, J.: P-Tuning v2: Prompt tuning can be comparable to fine-tuning universally across scales and tasks. In: ACL (2022)

\bibitem{houlsby2019adapter}
Houlsby, N., Giurgiu, A., Jastrzebski, S., Morrone, B., De Laroussilhe, Q., Gesmundo, A., Attariyan, M., Gelly, S.: Parameter-efficient transfer learning for NLP. In: ICML (2019)

\bibitem{zhang2023controlnet}
Zhang, L., Agrawala, M.: Adding conditional control to text-to-image diffusion models. arXiv preprint arXiv:2302.05543 (2023)

\bibitem{mou2023t2iadapter}
Mou, C., Wang, X., Xie, L., Wu, Y., Zhang, J., Qi, Z., Shan, Y., Qie, X.:
T2I-Adapter: Learning adapters to dig out more controllable ability for text-to-image diffusion models. arXiv preprint arXiv:2302.08453 (2023)

\bibitem{ye2023ipadapter}
Ye, H., Zhang, J., Liu, S., Han, X., Yang, W.: IP-Adapter: Text compatible image prompt adapter for text-to-image diffusion models. arXiv preprint arXiv:2308.06721 (2023)

\bibitem{gal2023textualinversion}
Gal, R., Alaluf, Y., Atzmon, Y., Patashnik, O., Bermano, A.H., Chechik, G., Cohen-Or, D.: An image is worth one word: Personalizing text-to-image generation using textual inversion. In: ICLR (2023)

\bibitem{mokady2023nulltext}
Mokady, R., Hertz, A., Bermano, A.H.: Null-text inversion for editing real images using guided diffusion models. In: CVPR (2023)

\bibitem{hessel2021clipscore}
Hessel, J., Holtzman, A., Forbes, M., Le Bras, R., Choi, Y.: CLIPScore: A reference-free evaluation metric for image captioning. In: EMNLP (2021)

\bibitem{hertz2023prompt2prompt}
Hertz, A., Mokady, R., Tenenbaum, J., Aberman, K., Pritch, Y., Cohen-Or, D.: Prompt-to-prompt image editing with cross-attention control. In: ICLR (2023)

\bibitem{ho2022cfg}
Ho, J., Salimans, T.: Classifier-free diffusion guidance. arXiv preprint arXiv:2207.12598 (2022)

\bibitem{tumanyan2023pnp}
Tumanyan, N., Geyer, M., Bagon, S., Dekel, T.: Plug-and-play diffusion features for text-driven image-to-image translation. In: CVPR, pp. 1921--1930 (2023)

\bibitem{ruiz2023dreambooth}
Ruiz, N., Li, Y., Jampani, V., Pritch, Y., Rubinstein, M., Aberman, K.: DreamBooth: Fine tuning text-to-image diffusion models for subject-driven generation. In: CVPR (2023)

\bibitem{hertz2024stylealigned}
Hertz, A., Voynov, A., Fruchter, S., Cohen-Or, D.: Style aligned image generation via shared attention. In: CVPR, pp. 4775--4785 (2024)

\bibitem{liu2022composable}
Liu, N., Li, S., Du, Y., Torralba, A., Tenenbaum, J.B.: Compositional visual generation with composable diffusion models. In: ECCV, pp. 423--439 (2022)

\end{thebibliography}
%

\end{document}